\documentclass{article}
\usepackage{spconf,amsmath,graphicx}
\usepackage{tabu, tabularx}

\usepackage{textcomp}
\usepackage{microtype}
\usepackage{hyperref}

\usepackage{booktabs,makecell,tabularx}

\newcolumntype{C}[1]{>{\centering\arraybackslash}p{#1}}
\newcolumntype{L}{>{\raggedright\arraybackslash}X}

\usepackage{siunitx}
\usepackage{etoolbox}
\newrobustcmd{\B}{\bfseries}

\title{Dyn-ASR: Compact, Multilingual Speech Recognition via Spoken Language and Accent Identification}
\name{Sangeeta Ghangam Manepalli$^{\star}$ \qquad Daniel Whitenack$^{\dagger}$ \qquad Joshua Nemecek$^{\dagger}$}

\address{$^{\star}$ Intel Corporation, Chandler, Arizona, USA \\
    $^{\dagger}$ SIL International, Dallas, Texas, USA}

\begin{document}
%\ninept
%
\maketitle
\begin{abstract}
Running automatic speech recognition (ASR) on edge devices is non-trivial due to resource constraints, especially in scenarios that require supporting multiple languages. We propose a new approach to enable multilingual speech recognition on edge devices. This approach uses both language identification and accent identification to select one of multiple monolingual ASR models on-the-fly, each fine-tuned for a particular accent. Initial results for both recognition performance and resource usage are promising with our approach using less than 1/12th of the memory consumed by other solutions.
\end{abstract}
\section{Introduction}
\label{sec:intro}
Speech-driven interfaces have become increasingly common, and COVID-19 has accelerated interest in speech as one of the primary modes of interaction with a variety of systems~\cite{bsound}. In the case of kiosks used for ticketing, banking or informational purposes, developers of speech interfaces need to minimize latency and resource consumption while maintaining a high level of performance in speech-related tasks, like automatic speech recognition (ASR). At the same time, kiosks placed in transportation or business hubs (e.g., airports or international hotels) need to be able to support multiple languages. 

Running any speech or language model on an edge system is non-trivial due to the size of parameter sets in modern speech and language models and the accelerated hardware needed to run many of these neural network based models. If multiple models are needed for multiple languages, resource consumption~\cite{asrmem} and increased inference times could very easily prevent developers from deploying applications in any environment, much less at the edge. In fact, common enterprise ASR systems require developers to deploy separate, dissimilar instances of model servers for each supported language, which complicates infrastructure and could result in reliability or maintenance issues. 

Beyond language, accent and other demographic factors have been shown to dramatically impact ASR performance~\cite{accent1, meyer-etal-2020-artie}. Those demographic factors need to be addressed via larger, more computationally expensive speech models or via models fine-tuned for particular demographic groups. 

In this paper, we implement and test an approach that allows for dynamic usage of monolingual, accent-specific speech recognition models in a multilingual context. This dynamic usage of monolingual models reduces the need for higher end compute and memory, preserves the recognition performance of monolingual ASR models, and ultimately enables multilingual contactless interactions on edge devices. In terms of user experience, the proposed methodology also removes the necessity for an awkward, touch-driven user selection of a language within an application. 

\section{Related Work}
\label{sec:relwork}

Many speech-driven systems at the edge detect a wake word and transfer audio to the cloud or another remote endpoint for processing. Such a system may not be desirable in cases where latency and privacy are a concern. For example, per Google Cloud Speech-to-Text best practice recommendations~\cite{gcloud}, 16-bit, 16kHz mono PCM audio data needs to be sent to API endpoints with a frame size of 100ms. Thus, round trip latency would be around 4s for an upload of around 3.2KB/second, which is beyond acceptable levels for many applications.

In terms of language support, developers often leverage slot-filling. For example, developers might require a user to select a preferred language via a touchscreen or speak the name of their preferred language. Both of these interactions could be awkward as they represent interactions that do not naturally occur in conversation. Moreover, the former implementation prevents truly contactless interactions, and the latter presents challenges in dealing with alternate language names, accents, and demographic factors.

Researchers have tried to extend ASR systems that rely on phonetically inspired acoustic models to support multiple languages. Certain of these extensions pool phonemes from all languages into a single set and others manage separate sets (see~\cite{Li2020UniversalPR} and references therein). Either approach requires the management of expertly crafted linguistic information, and the burden of curating that information grows with the number of languages. 

There have also been an increasing number of attempts to create end-to-end (E2E) models that recognize speech in multiple languages. For example, \cite{Pratap2020MassivelyMA} utilizes a single model which is trained while sharing parameters across 51 languages, and the model ends up containing over a billion parameters. The scale of this type of model would make it a challenge to utilize on edge devices because of resource consumption and/or the need for specialized hardware accelerators. Other attempts at E2E multilingual ASR exhibit similar characteristics~\cite{Watanabe2017LanguageIE}.

In~\cite{Kannan2019LargeScaleMS}, the authors use an E2E architecture called RNN Transducer (RNN-T) that shows promise in edge applications. A larger, adapted version of RNN-T takes as input a vector representing a particular one of 9 supported languages. The authors assume that the language is either specified manually by a user of the model or determined automatically from a language identification system, but they do not integrate any specific spoken language identification system.  

In still another vein of research~\cite{Seki2018AnEL, Punjabi2020StreamingEB, Weiner2012IntegrationOL, Li2019TowardsCA, Caesar2012INTEGRATINGLI, GonzalezDominguez2015ARE}, researchers have tried to integrate spoken language identification directly into a joint speech recognition and language ID model. Generally, this work adds the prediction of one or more language identifications into the prediction of other outputs, such as decoded characters. While these models can reduce the overall number of models needed to support multilingual ASR, they can exhibit degradation in performance for some or all of the supported languages in comparison to the performance of corresponding monolinguhttps://www.overleaf.com/project/6020b5d855cbeaf4a4bfea22al ASR models. Moreover, the authors are not aware of any of these studies that integrates identifications of accent intohttps://www.overleaf.com/project/6020b5d855cbeaf4a4bfea22f speech recognition models. 

While the present study is related to recent work in spoken language identification~\cite{Aaltodoc:http://urn.fi/URN:NBN:fi:aalto-202006213891} and E2E ASR, it capitalizes on a new, simultaneous combination of spoken language identification, spoken accent identification, and fine-tuned E2E ASR, which was not considered in these earlier studies. In particular, we investigate the efficacy of utilizing such a combination on edge devices for high quality multilingual ASR.

\section{Approach}
\label{sec:approach}

\begin{figure}[ht]
\centering
\includegraphics[width=0.38\textwidth]{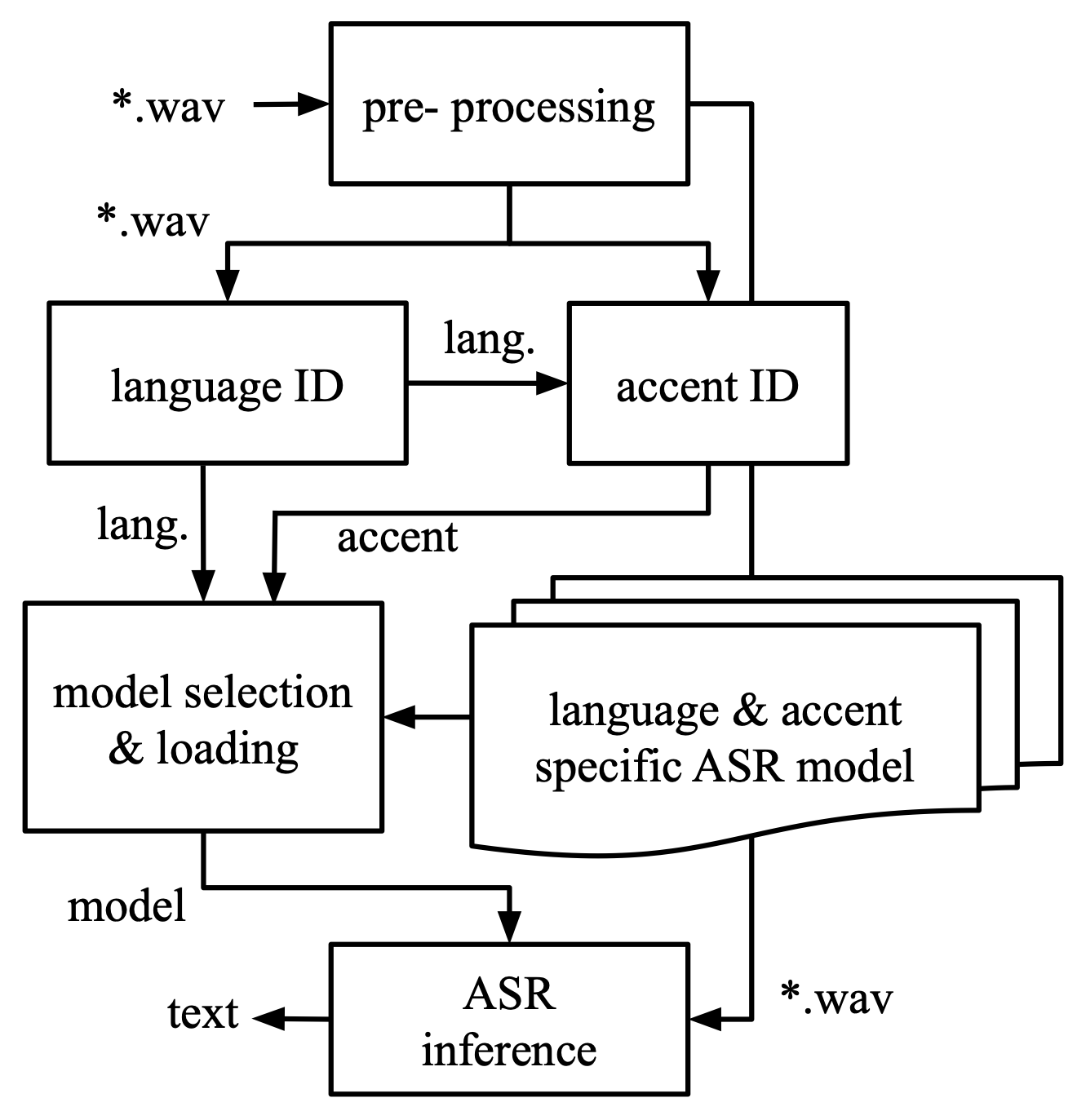}
\caption{The proposed processing steps and flow for recognizing multiple languages.}
\label{fig:cflow}
\end{figure}

An overview of the proposed Dynamic ASR (or Dyn-ASR, which we pronounce "dinosaur") to processing multilingual speech is presented in Fig.~\ref{fig:cflow}. In the approach, we assume WAV file inputs which are first pre-processed to normalize, trim silence from, and format the audio. In the case of audio fed to speech recognition models, the audio is formatted to 16-bits and 16kHz. In case of audio fed to language identification and accent identification models, the audio is formatted to 16-bits and 8kHz, and we artificially repeat the input audio to fill at least 10 seconds.  

After pre-processing, language and accent identification is performed. For both language and accent identification we utilize a model with two LSTM layers, each having 200 units and each followed by batch normalization. One such model is used to classify the input audio into a language class. Then, we utilize a separate accent identification model (corresponding to the identified language) to further classify the input audio into an accent class. 

An ASR model is trained for each language and accent pair that is to be supported by the system. We fine-tune these language and accent specific ASR models from general (i.e., not accent specific) ASR models. Because we are targeting edge applications, we experimented with several different phonetically inspired and E2E model architectures that were optimized for edge devices using OpenVINO~\cite{Gorbachev2019OpenVINODL} and/or compact by nature. These models included Deep Neural Network (DNN) acoustic models and RNN-T, Conformer~\cite{Gulati2020ConformerCT}, DeepSpeech~\cite{Hannun2014DeepSS}, and QuartzNet~\cite{Kriman2020QuartznetDA} E2E models. In the end, we found that the Conformer and/or QuartzNet E2E models fulfilled our constraints in terms of ASR performance and system resource consumption. 

Depending on the size of the models and system constraints, each of the ASR models can be loaded into memory when an application implementing the Fig.~\ref{fig:cflow} process starts, or each ASR model could be loaded into memory on-the-fly. In any event, the model corresponding to the identified language and accent pair is dynamically chosen or loaded into memory after the language and accent is identified. In this way, multiple compact monolingual models can be utilized dynamically to recognize speech in multiple languages without significantly sacrificing the performance of speech recognition or exceeding edge device memory or processor constraints.

\section{Experiments}
\label{sec:experiments}

To test the Fig.~\ref{fig:cflow} process for multilingual speech recognition on edge devices, we evaluated (i) the performance of our language and accent identification models; (ii) the performance of our language and accent specific ASR models; and (iii) the performance of an implementation of the full Fig.~\ref{fig:cflow} process with respect to resource consumption.

\subsection{Data}
\label{ssec:data}

In the following, we trained and tested our models/methods on English, Tamil, and Mandarin speech data. The English data was segmented into 8 accents (Scotland, Australia, England, India, USA, China, Malaysia, other) and the Mandarin data was segmented into 3 accents (Mainland, Taiwan, Hong Kong).

For transcribed speech data with corresponding language and accent labels, we relied on Mozilla's Common Voice data, the Speech Accent Archive from George Mason University (SAA), and the Singapore National Speech Corpus (NSC). For additional Tamil speech data, we used Microsoft's Indian Language Speech Corpus. We used the SoX utility to normalize the speech files to 16kHz, 16-bit WAV files for training and testing ASR tasks and 8kHz, 16-bit WAV files for training and testing language and accent identification tasks. 

\subsection{LID and accent identification}
\label{ssec:lid}

As mentioned in Section~\ref{sec:approach}, we utilize one LSTM-based model for language identification and one LSTM-based model per language for accent identification. For our combination of English, Tamil, and Mandarin, that means that we have 1 spoken language identification model and 2 spoken accent identification models (one for English and one for Mandarin). We sampled 38,400 samples per language to train the models. To train the accent identification models we utilized rejection sampling due to the unbalanced nature of the accent data.  

Our language identification model gives 84.99~\% accuracy on the 3 language classes. On English accents, we achieve 74.41~\% accuracy across the 8 accents, and we achieve 79.83~\% accuracy across the 3 Mandarin accents. The more crucial step in the Fig.~\ref{fig:cflow} approach is language identification, because language identification determines if the ASR model used will correspond to the spoken language or another language entirely. Correct accent identification will further improve recognition accuracy, but to a lesser degree. Our results here show that executing a spoken language identification model prior to selection of ASR model could result in choosing a proper model for at least 8-9 out of 10 inferences. Additionally, we found that using a single model for LID and accent identification would not achieve comparable accuracy on a similarly-sized model. Using a larger model for combined LID and accent identification would also slow down the time-to-ASR for the combined system.

\subsection{Fine-tuned ASR}
\label{ssec:dynasr}

Assuming a proper language identification, we also wanted to validate the idea that switching between monolingual ASR models (each fine-tuned for a particular accent) could both: (i) outperform individual models trained on data corresponding to multiple accents; and (ii) allow us to avoid more complicated and/or larger multi-accent data and models. We created a test set of Indian, Chinese, and Malaysian accented English by selecting these accents out of the SAA. We then evaluated ASR models fine-tuned on each of these accents alongside publicly available pre-trained models. For this evaluation, we chose English because of the availability of multiple pre-trained models for comparison and because it is one of the languages considered in our other experiments.

The ASR models we fine-tuned were based on the QuartzNet architecture and fine-tuned on Indian, Chinese, and Malaysian accented English data from the Singapore National Speech Corpus. When evaluating these models (collectively referred to below as the models of the Dynamic ASR system, or Dyn-ASR), we loaded and utilized each of the models for the corresponding annotated accent. This simulates the best case scenario when loading language and accent specific models in the process illustrated in Fig.~\ref{fig:cflow}. Of course in any implementation of the Fig.~\ref{fig:cflow}, the performance of the Dyn-ASR models will depend on the performance of the language and accent identification models, but this evaluation gives us a baseline for evaluating the set of ASR models themselves.

The pre-trained models that we used as a reference are DeepSpeech trained on US English (DS), QuartzNet trained on LibriSpeech (QN-LS), and QuartzNet trained on multiple accents (QN-Multi). The results of this comparison are presented in Table~\ref{table:asr}. 

\begin{table}[!htb]
    \caption{Accent Specific ASR Results. All results are word error rates (WERs) per accent and per model or set of models. }
    \vspace{10pt}
    \label{table:asr}
    \sisetup{table-format=1.0}% table-column-width =2.5cm
    \centering
    \small
    \begin{tabular} {l | l l l l } 
        \toprule
{\thead{Model(s):}}  
    & {\thead{DS} } 
        & {\thead{QN-LS}}
            & {\thead{QN-Multi}}
                & {\thead{Dyn-ASR}}\\
    \midrule
India & 45.93 & 32.15 & 16.02 & \textbf{15.89} \\
China  & 57.10 &  44.01 &  27.45 & \textbf{26.05} \\
Malaysia  & 42.75 & 19.75 &  13.59 &  \textbf{11.59} \\
    \bottomrule
    \end{tabular}
\end{table}

\subsection{Resource consumption}
\label{ssec:resource}

\begin{figure}[ht!]

\begin{minipage}[b]{1.0\linewidth}
  \centering
  \centerline{\includegraphics[width=8.5cm]{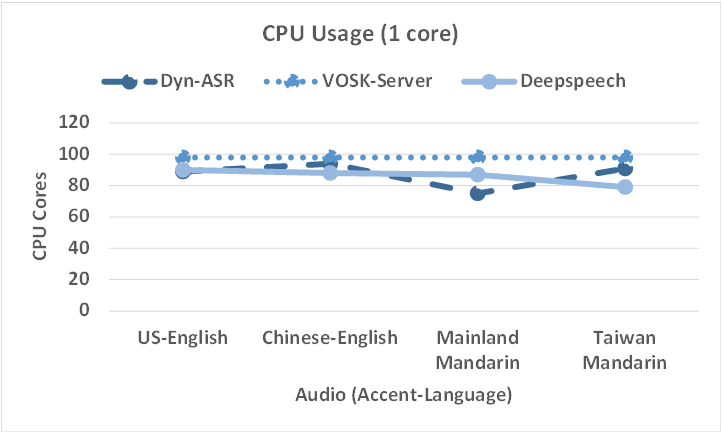}}
%  \vspace{2.0cm}
  \centerline{(a) CPU Usage}\medskip
\end{minipage}
\begin{minipage}[b]{1.0\linewidth}
  \centering
  \centerline{\includegraphics[width=8.5cm]{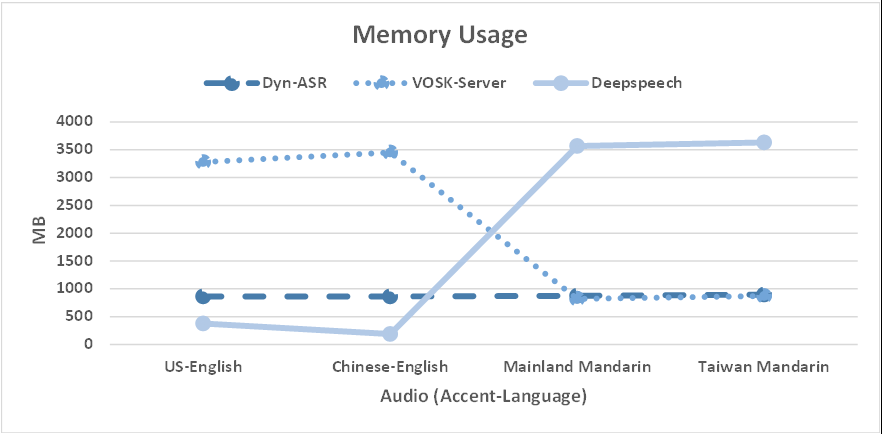}}
%  \vspace{1.5cm}
  \centerline{(b) Memory Usage}\medskip
\end{minipage}
%\hfill
\begin{minipage}[b]{1.0\linewidth}
  \centering
  \centerline{\includegraphics[width=8.5cm]{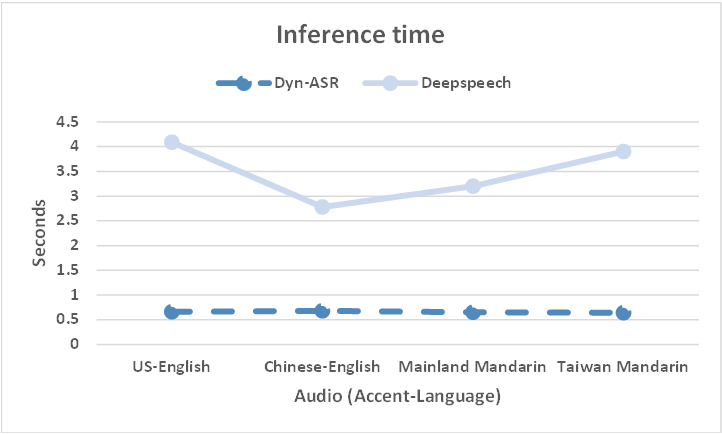}}
%  \vspace{1.5cm}
  \centerline{(c) Inference Time}\medskip
\end{minipage}
\caption{Resource usage results for Dyn-ASR, VOSK-Server (VS-EN, VS-CMN) and Deepspech (MDS-EN, PDS-CMN). The Dyn-ASR solution uses minimal memory and has lowest inference time}
\label{fig:res}
\end{figure}

To evaluate resource consumption, we created an implementation of the Fig.~\ref{fig:cflow} approach for English and Mandarin. We compare the resource consumption of this implementation (Dyn-ASR below) with the Vosk speech recognition toolkit server~\cite{vosk} (both an English instance, VS-EN, and a Mandarin instance, VS-CMN), Mozilla's DeepSpeech implementation trained on US English (MDS-EN)~\cite{Hannun2014DeepSS}, and PaddlePaddle's DeepSpeech implementation trained on Mandarin (PDS-CMN). Note, the authors had difficulty in finding any practical, publicly available system natively supporting multilingual ASR models or integrating spoken language identification. As such, multiple instances and versions of these systems had to be deployed, which demonstrates the operational barriers to practically deploying a multilingual ASR system. 

While there are portable versions of the Vosk servers for each language, we picked the server version that would give the best quality speech recognition  results. We utilized two languages (English and Mandarin), each with two accents (US and Chinese accented English and Mainland and Taiwanese accented Mandarin) respectively for the input audio data. Table~\ref{table:resource} includes the system resource consumption for each  solution.

\begin{table}[!htb]
    \caption{Storage and Memory in MB used by each solution. Dyn-ASR utilizes only 10~\% and 25~\% of highest storage and memory consumed by the other solutions. DeepSpeech models are not loaded in memory at installation.}
    \vspace{10pt}
    \label{table:resource}
    \sisetup{table-format=1.5}% table-column-width =2.5cm
    \centering
    \small
    \begin{tabular} {l | l l l l l} 
        \toprule
\thead {System}  
    & {\thead{Dyn-ASR} } 
        & {\thead{VS-EN}}
            & {\thead{VS-CMN}}
                & {\thead{MDS- \\ EN}}
                    & {\thead{PDS- \\CMN}}\\
    \midrule
Storage & 317 &  3200 & 426 &  1000 & 2800\\
Memory  & 775 &  3049 & 469 &  NA & 3130\\
    \bottomrule
    \end{tabular}
\end{table}

To ensure that we could provide each solution with whatever resources it could consume, we ran all of the ASR solutions on a Core i9 System (i9-7920X CPU) which has 12 cores and 24 threads with 64GB of system memory and 500GB of storage. All the audio file inputs were of type 16kHz, 16-bit PCM mono. Results were captured in terms of the number of CPU cores utilized, memory usage and total inference time and are included in Fig.~\ref{fig:res}

As shown in Fig.~\ref{fig:res} part \textit{a}, the Dyn-ASR container has not been pinned to a CPU core, and thus it ends up using as many cores as needed to complete inference at a constant time of less than 1 second (see Fig.~\ref{fig:res} part \textit{c}). It also uses minimal incremental memory as shown in Fig.~\ref{fig:res} part \textit{b} (around 10MB). A combination of the VS-EN + VS-CMN systems or the MDS-EN + PDS-CMN systems would need to be assembled to match the multilingual ASR capabilities of Dyn-ASR, yet any of these combinations would exceed the memory usage of the the example Dyn-ASR system and increase the complexity of deployed infrastructure. Further, neither of these combinations (VS-EN + VS-CMN or MDS-EN + PDS-CMN) would solve the problem of selecting the correct ASR model corresponding to the input language, which functionality is natively rolled into the Dyn-ASR system. These characteristics together make the Dyn-ASR approach appealing for edge deployments.

\section{Conclusions and Future Work}
\label{sec:conclusion}
We introduced a new approach to multilingual speech recognition that selectively uses monolingual ASR models fine-tuned for particular accents. The particular recognition models used for each inference is determined on-the-fly using a language identification model and an accent identification model. An implementation of this approach for English and Mandarin behaved favorably in terms of resource consumption as compared to other publicly available ASR solutions and also shows promise in terms of recognition performance. This work explored certain model architectures, but we are exploring still other architectures along with further optimization using Intel's OpenVINO toolkit. The authors would also like to integrate a step in the processing that uses a text-based model and/or probabilities from the language/ accent identification to deal with misidentified languages. 

\vfill\pagebreak

\bibliographystyle{IEEEtran}
\bibliography{mds}{}
\end{document}